\documentclass[conference,a4paper]{APSIPA2021}
\usepackage{multirow}
\usepackage{graphicx}
\usepackage{amsmath}
\usepackage{amssymb}
\usepackage{amsxtra}
\usepackage{threeparttable}
\graphicspath{{figs/}}
\usepackage{booktabs}
\usepackage{etoolbox}
\usepackage{siunitx}
\usepackage{cite}
\usepackage{algorithm,algpseudocode}
\usepackage[caption=false, font=footnotesize]{subfig} 

\algnewcommand{\algorithmicforeach}{\textbf{for each}}
\algdef{SE}[FOR]{ForEach}{EndForEach}[1]
  {\algorithmicforeach\ #1\ \algorithmicdo}
  {\algorithmicend\ \algorithmicforeach}

\begin{document}

\title{StyleGAN Encoder-Based Attack for Block Scrambled Face Images}

\author{%
\authorblockN{%
AprilPyone MaungMaung and
Hitoshi Kiya
}
\authorblockA{%
Tokyo Metropolitan University, Tokyo, Japan
}
%
%
}

\maketitle
\thispagestyle{empty}

\begin{abstract}
  In this paper, we propose an attack method to block scrambled face images, particularly Encryption-then-Compression (EtC) applied images by utilizing the existing powerful StyleGAN encoder and decoder for the first time.
  Instead of reconstructing identical images as plain ones from encrypted images, we focus on recovering styles that can reveal identifiable information from the encrypted images.
  The proposed method trains an encoder by using plain and encrypted image pairs with a particular training strategy.
  While state-of-the-art attack methods cannot recover any perceptual information from EtC images, the proposed method discloses personally identifiable information such as hair color, skin color, eyeglasses, gender, etc.
  Experiments were carried out on the CelebA dataset, and results show that reconstructed images have some perceptual similarities compared to plain images.
\end{abstract}

\section{Introduction}
Nowadays, anyone with a smartphone can generate multimedia data (image, video and audio).
Moreover, all sorts of cameras such as webcams, dashcams, security cameras, etc.\ are capturing images and videos every moment.
Sheer amount of generated multimedia data are processed by distributed systems.
Although we can enjoy many distributed multimedia applications, security and privacy issues are at an alarming high rate.
Secure and efficient communication of multimedia data require both compression and encryption~\cite{zhou2013designing,chuman2019encryption}.
For encryption, full encryption with provable security (such as RSA and AES) is the most secure option.
However, many multimedia applications have been seeking a trade-off between security and other features such as low processing demands, tolerance of data loss, bitstream compliance, compatibility with machine learning, and signal processing in the encrypted domain.
Therefore, perceptual encryption methods have been studied and developed to balance the trade-off~\cite{wang2020color,tang2015efficient,ito2009one,kurihara2015encryption,chuman2019encryption}.

In the context of secure image transmission, the traditional way is to use a Compression-then-Encryption (CtE) system.
However, social networking services (SNS) and cloud photo sharing services (CPSS) require to (multiply) re-compress uploaded images.
To be in line with such services, it is preferred to use an Encryption-then-Compression (EtC) system~\cite{chuman2019encryption}.
In addition, a recent study shows that EtC images can be classified with an isotropic network~\cite{maung2022privacy}.
Therefore, to reliably use EtC images in many applications, security evaluation has become important.

\begin{figure}[!t]
\centering
\subfloat[]{\includegraphics[width=0.2\linewidth]{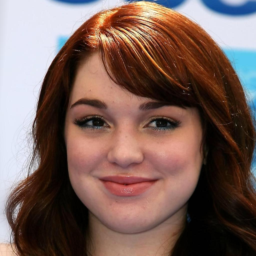}%
\label{fig:plain}}
\hfil
\subfloat[]{\includegraphics[width=0.2\linewidth]{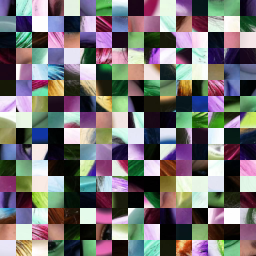}%
\label{fig:encrypted}}
\hfil
\subfloat[]{\includegraphics[width=0.2\linewidth]{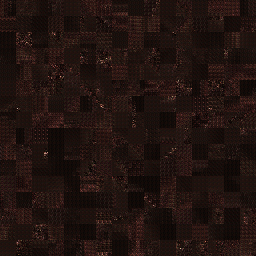}%
\label{fig:itn}}
\hfil
\subfloat[]{\includegraphics[width=0.2\linewidth]{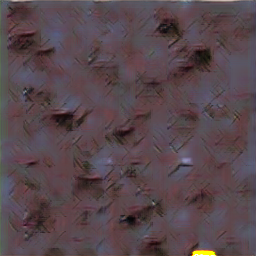}%
\label{fig:gan}}
\hfil
\subfloat[]{\includegraphics[width=0.2\linewidth]{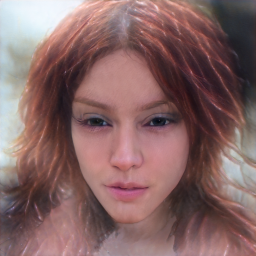}%
\label{fig:rec}}
\caption{Example of reconstructed image from different attacks. (a) Plain. (b) Encrypted. (c) ITN~\cite{ito2021image}. (d) GAN~\cite{madono2021gan}. (e) Proposed. The proposed attack reveals hair color and gender of encrypted image in this example.\label{fig:recon-examples}}
\end{figure}

The previous attacks exploit the correlation between pixels~\cite{chang2020attacks,chuman2017security}, or deep neural networks~\cite{madono2021gan,ito2021image} to recover visual information from encrypted images.
EtC images are robust against such attacks.
In this paper, we argue that it is possible to recover some visual information to disclose an individual's identity with the help of existing generative models.
Figure~\ref{fig:recon-examples} shows examples of reconstructed images from the previous attacks and the proposed one.

For the first time, we propose an attack method that utilizes a StyleGAN encoder proposed in~\cite{richardson2021encoding} with a particular training strategy to reconstruct personally identifiable visual information from EtC images.
The proposed method extracts styles from the EtC images and injects the extracted styles into the pre-trained StyleGAN2 model~\cite{karras2020analyzing}.
We make the following contributions in this paper.
\begin{itemize}
  \item By utilizing the available StyleGAN encoder and decoder, we propose an attack method in which we deploy a particular training strategy for training the encoder to extract the styles from encrypted images.
  \item We conduct experiments on a face dataset and validate results with a perceptual information measure.
\end{itemize}
In experiments, the proposed attack method is confirmed to outperform the previous attack methods to attack privacy on the face dataset.

\section{Related Work}
\subsection{Block Scrambling-Based Image Encryption}
Block scrambling-based image encryption methods for EtC systems can withstand compression methods used by cloud service providers such as SNS and CPSS.\@
There are two kinds of such encryption: color-based image encryption~\cite{kurihara2015encryption,watanabe2015encryption,kurihara2015encryption2,kurihara2017encryption} and grayscale-based image encryption~\cite{chuman2019encryption,sirichotedumrong2019grayscale}.
In this paper, we focus on color-based image encryption which is compatible with the JPEG standard~\cite{kurihara2015encryption} and can also be used for deep neural network-based image classification~\cite{maung2022privacy}.

\begin{figure}[!t]
\centering
\includegraphics[width=70mm]{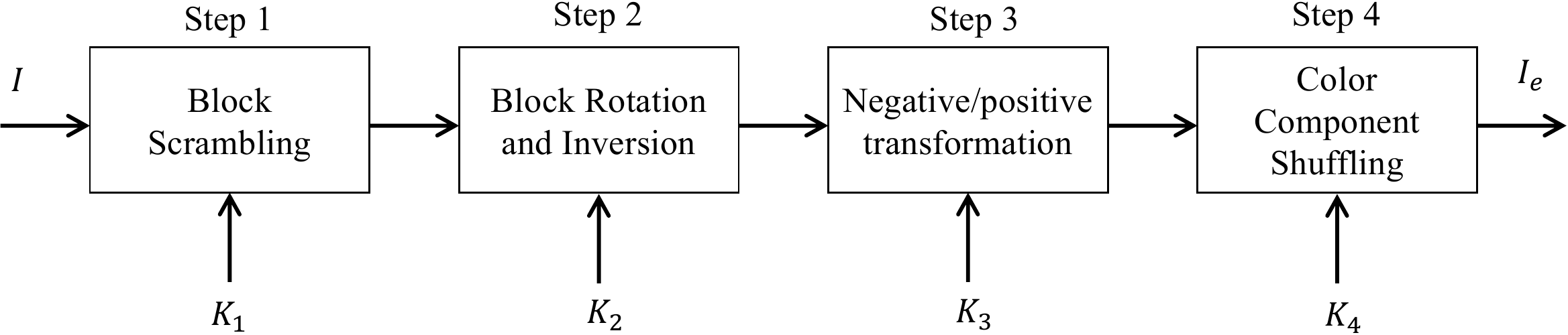}
\caption{Color-based image encryption steps for EtC.\label{fig:etc}}
\end{figure}

Figure~\ref{fig:etc} depicts the encryption steps of the color-based image encryption. A three-channel (RGB) color image ($I$) with $X \times Y$ pixels is divided into non-overlapping blocks each with $B_x \times B_y$. Then, four encryption steps are carried out on the divided blocks as follows.
\begin{enumerate}
 \item Randomly permute the divided blocks by using a random integer generated by a secret key $K_1$.
 \item Rotate and invert each block randomly by using a random integer generated by a key $K_2$.
 \item Apply negative/positive transformation to each block by using a random binary integer generated by a key $K_3$, where $K_3$ is commonly used for all color components. A transformed pixel value in the $i$\textsuperscript{th} block, $p'$, is calculated using
 \begin{equation}
 p' = \left\{
 \begin{array}{ll}
 p & (r(i) = 0)\\
 p \oplus (2^L - 1) & (r(i) = 1),
 \end{array}
 \right.
 \end{equation}
 where $r(i)$ is a random binary integer generated by $K_3$, $p$ is the pixel value of the original image with $L$ bits per pixel ($L=8$ is used in this paper), and $\oplus$ is the bitwise exclusive-or operation. The value of the occurrence probability $\mathrm{P}(r(i)) = 0.5$ is used to invert bits randomly.
 \item Shuffle three color components in each block by using an integer randomly selected from six integers generated by a key $K_4$.
\end{enumerate}

\begin{figure*}[!t]
\centering
\subfloat[]{\includegraphics[width=0.15\linewidth]{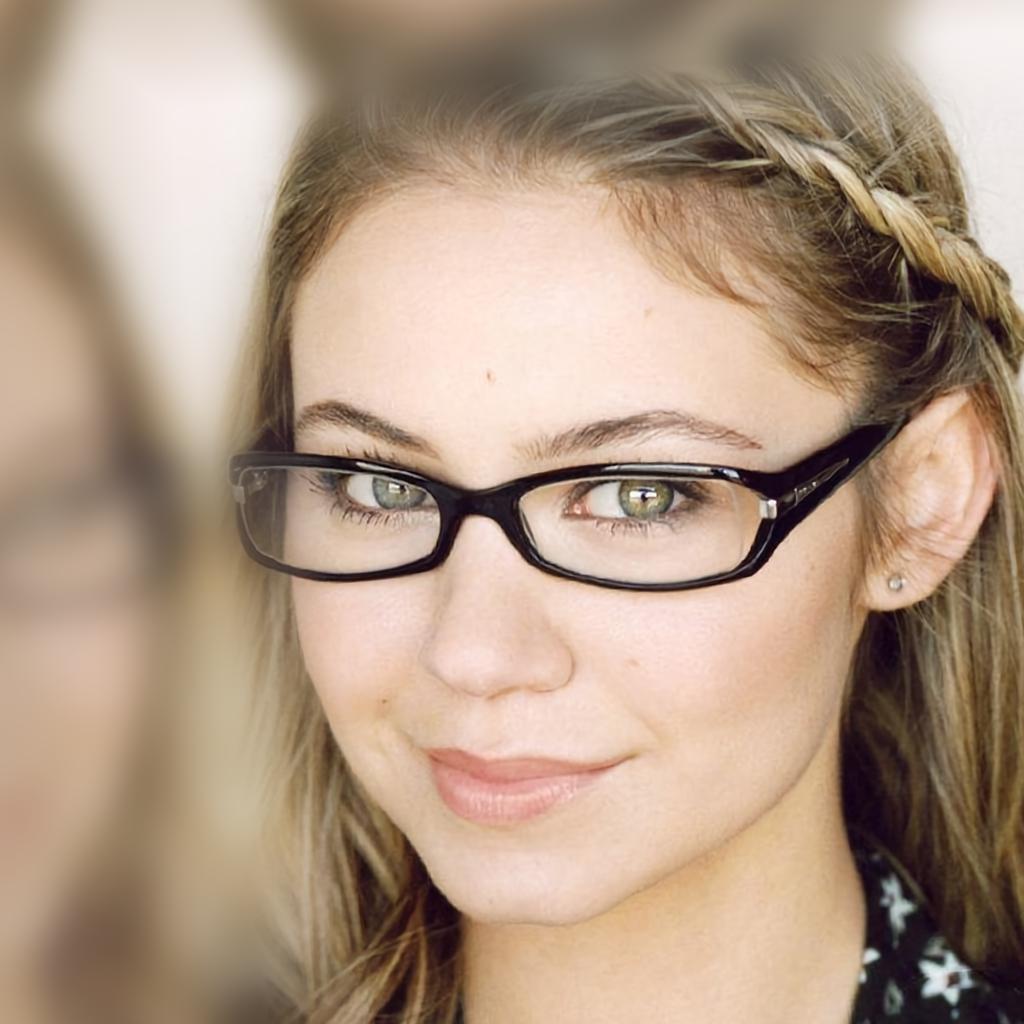}%
\label{fig:cele}}
\hfil
\subfloat[]{\includegraphics[width=0.15\linewidth]{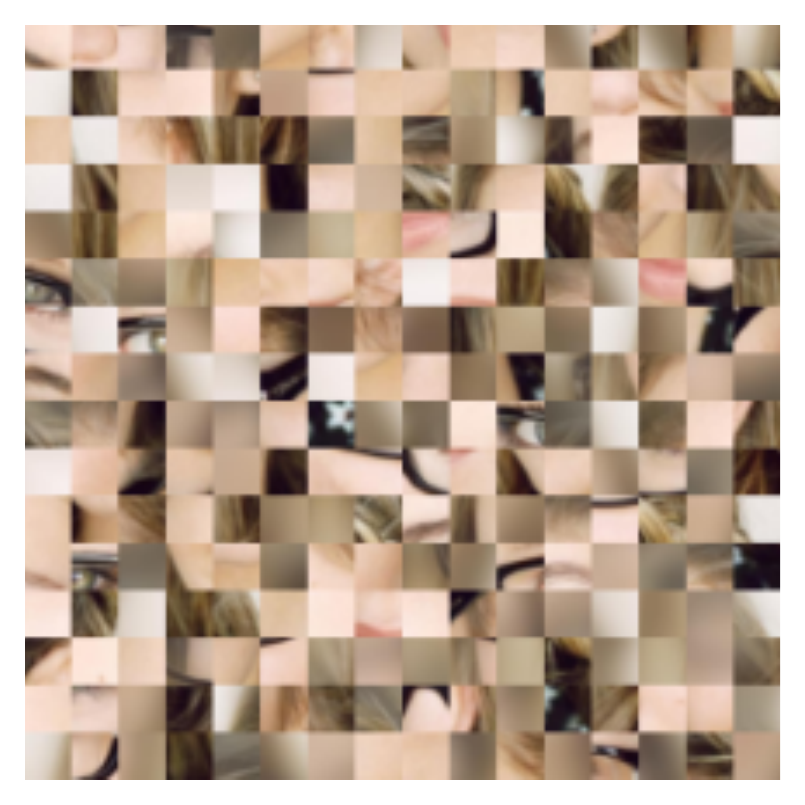}%
\label{fig:step-1}}
\hfil
\subfloat[]{\includegraphics[width=0.15\linewidth]{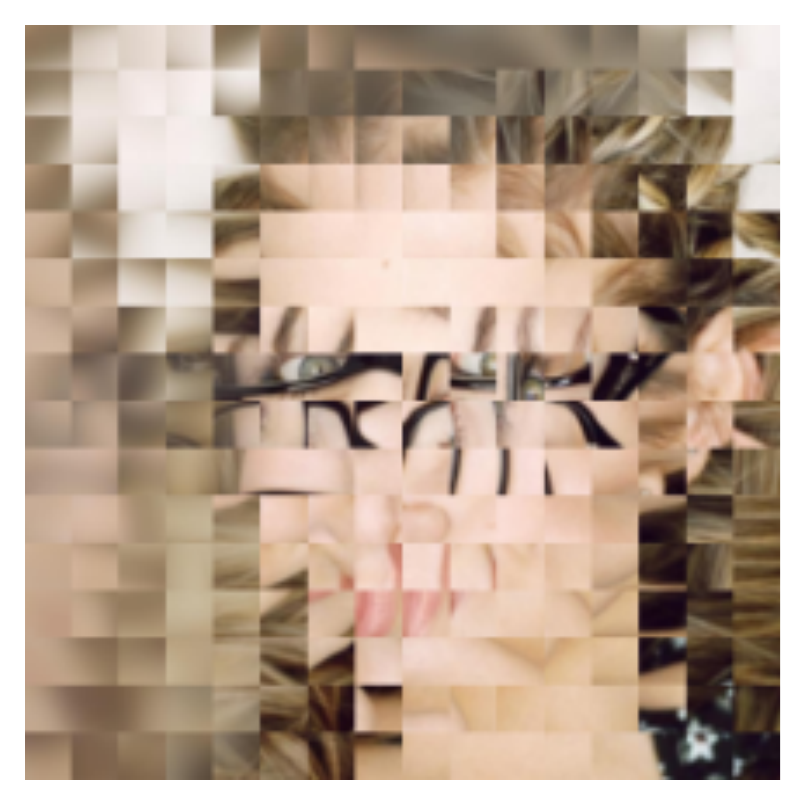}%
\label{fig:step-2}}
\hfil
\subfloat[]{\includegraphics[width=0.15\linewidth]{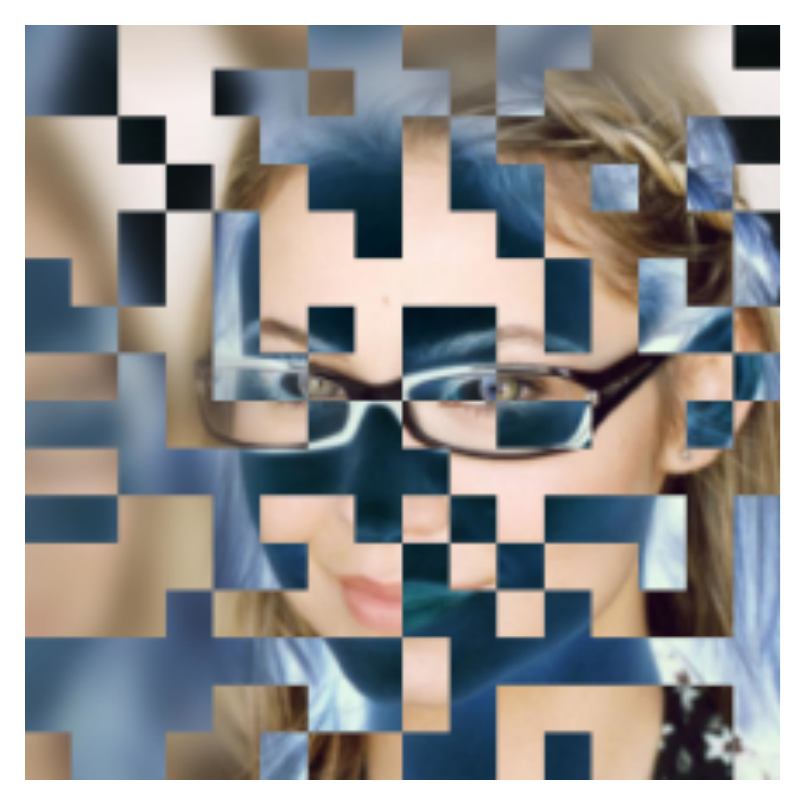}%
\label{fig:step-3}}
\hfil
\subfloat[]{\includegraphics[width=0.15\linewidth]{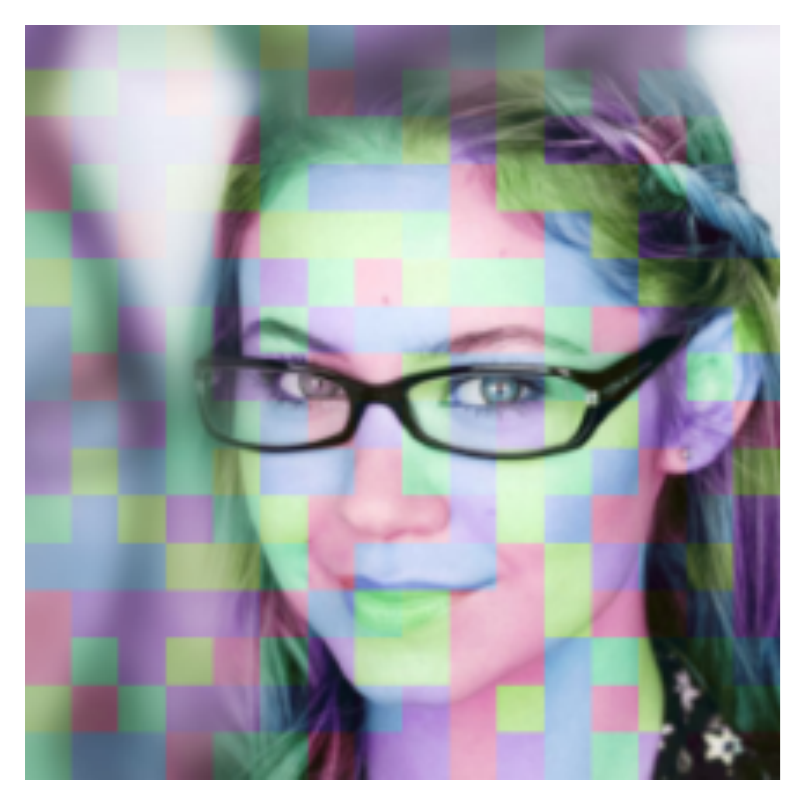}%
\label{fig:step-4}}
\hfil
\subfloat[]{\includegraphics[width=0.15\linewidth]{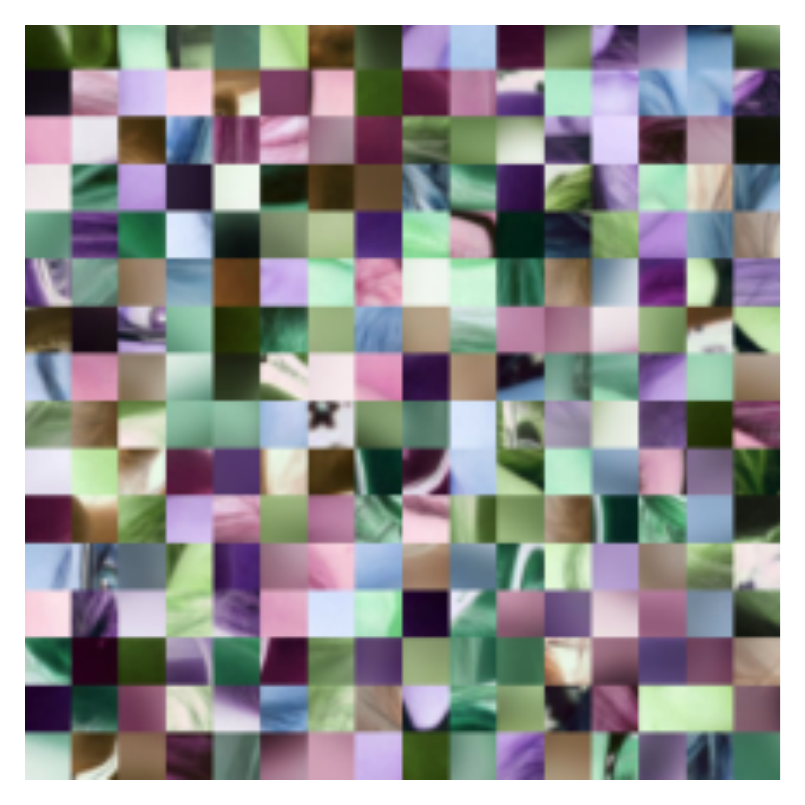}%
\label{fig:enc_cele}}
\caption{Example of encrypted image by each encryption step. (a) Plain. (b) Step 1 only. (c) Step 2 only. (c) Step 3 only. (d) Step 4 only. (e) All four encryption steps.\label{fig:etc-steps}}
\end{figure*}

Then, integrate the encrypted blocks to form an encrypted image $I_e$. Note that block size $B_x = B_y = 16$ is enforced to be JPEG compatible.
For visualization, Fig.~\ref{fig:etc-steps} shows the distortion of each encryption step.

\subsection{Brute-Force Attack}
Generally, to be robust against brute-force attacks, the key space has to be large enough.
The key space of the above color-based image encryption depends on the number of blocks, $n$ which is given by
\begin{equation}
 n=\biggl\lfloor \dfrac{X}{B_x}\times \dfrac{Y}{B_y} \biggr\rfloor,
\end{equation}
where $\lfloor \cdot \rfloor$ is a function that rounds down to the nearest integer, and $B_x$ and $B_y$ are the width and height of a block.
Therefore, the key space of the color-based encryption $\mathcal{K}(n)$ is given as below,
\begin{equation}
  \mathcal{K}(n)=n!\times8^n\times2^n\times6^n.
\end{equation}
For example, for an image size $(X = 256, Y = 256)$ with a block size $(B_{x}=16, B_{y}=16)$, there are $n = 256$ blocks.
In this case, the key space $\mathcal{K}(256)$ is much larger than a 256-bit key.
Therefore, the key space of the color-based encryption is generally large enough against the brute-force attacks.

Despite the large key space, an encrypted image has almost the same correlation among pixels in each block as that of the original image, whose property enables to efficiently compress images.
Therefore, an attacker might exploit this property to decrypt encrypted images.
We review such an attack in the following subsection.

\subsection{Jigsaw Puzzle Solver Attack}
Jigsaw Puzzle solver is a method of automatically assembling a Jigsaw Puzzle.
Blocks in block-based encryption are regarded as pieces of a Jigsaw Puzzle and decrypting an encrypted image is similar to assembling a Jigsaw Puzzle.
Therefore, the previous study considered Jigsaw Puzzle solvers as one of the attack methods to block scrambling-based image encryption~\cite{chuman2017icassp,chuman2017icme,chuman2018security}.
These solver attack methods utilize pairwise compatibility and pairwise comparison, and confirmed that assembling encrypted images was difficult if the number of blocks is large, the block size is small, and encrypted images have compression distortion and less color information.
Therefore, the color-based encryption method is also generally robust against traditional Jigsaw Puzzle solver methods.

\subsection{Inverse Transformation Attack}
This attack method was originally proposed to evaluate the transformation network-based encryption~\cite{ito2021image}.
However, it can also be considered as an attack method for block scrambling-based encryption methods~\cite{kiya2022overview}.
The main idea is to train a deep convolutional neural network to decrypt the encrypted images by using plain and encrypted image pairs.
This attack is possible because an attacker can prepare a synthetic dataset to train an inverse transformation network.
Although this method works for some encryption method~\cite{sirichotedumrong2021gan}, it is confirmed that the color-based image encryption is also robust against such attack methods~\cite{kiya2022overview}.

\subsection{GAN Attack}
Similarly, another recent neural network-based attack utilizes a generative adversarial network (GAN)~\cite{madono2021gan}.
This GAN-based method aims to reconstruct visual information by using the hinge loss.
Although the GAN-based attack was successfully applied to learnable image encryption methods such as~\cite{2018-ICCETW-Tanaka,2019-Access-Warit}, it could not recover visual information from EtC images~\cite{madono2021gan}.
Therefore, the color-based image encryption is also robust against the conventional GAN-based attack because the color-based encryption contains a block scrambling encryption step.

This paper aims to show that it is possible to recover visual information from encrypted images so that the security of color-based encrypted images for EtC systems can be improved in the future.

\section{Proposed Attack}
\subsection{Overview}
\begin{figure}[!t]
\centering
\includegraphics[width=\linewidth]{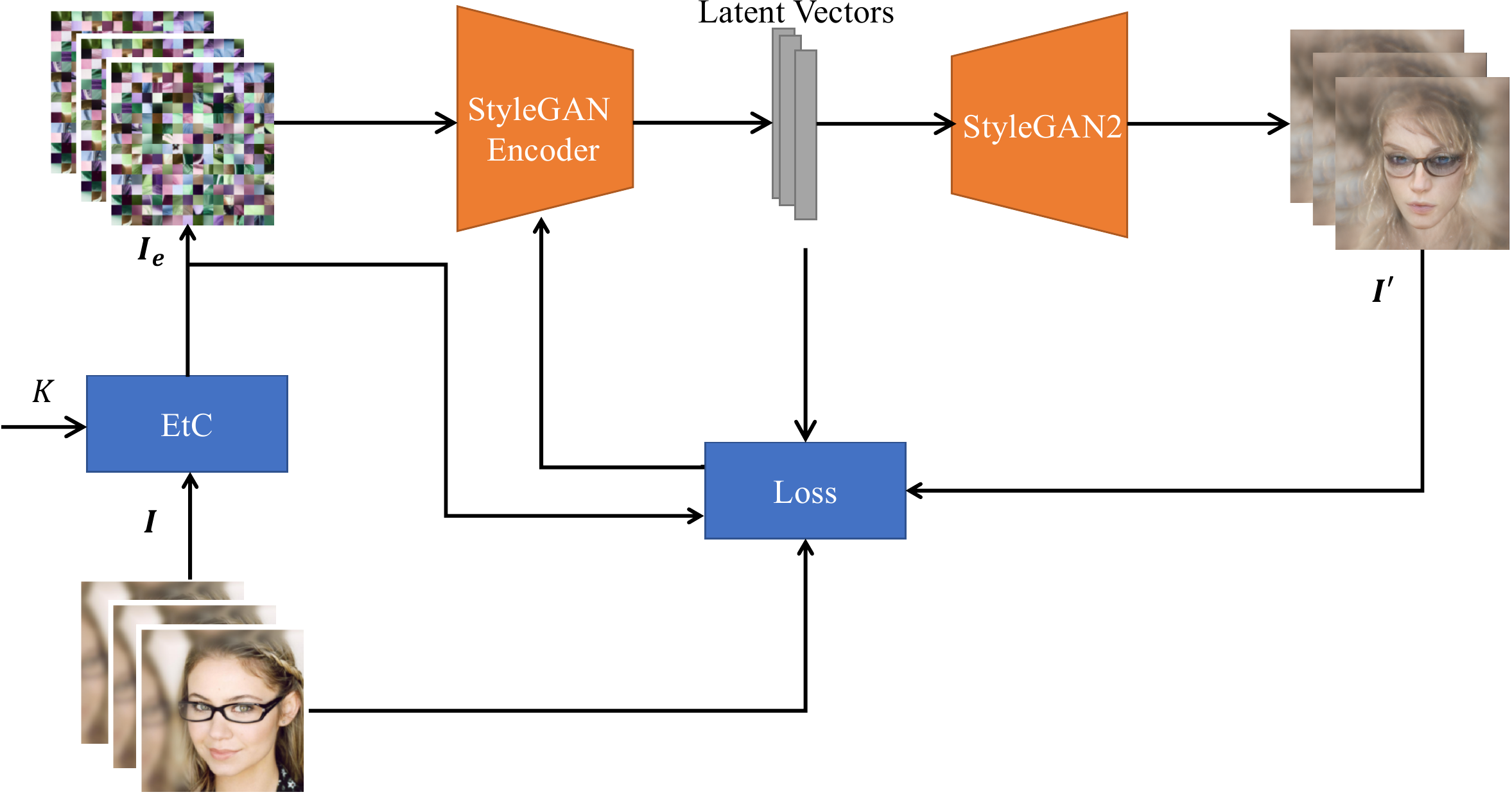}
\caption{Overview of proposed attack.\label{fig:etc-attack}}
\end{figure}

Figure~\ref{fig:etc-attack} shows an overview of the proposed attack.
Without loss of generality, we use $I$ for a plain image, $I'$ for a reconstructed image, and $I_e$ for an encrypted image, and their bold counterparts ($\boldsymbol{I},\boldsymbol{I'}, \boldsymbol{I_e}$) refer to images in a batch.
The main idea is to extract styles from an encrypted image and the extracted styles are used to generate an image similar to the original one to reveal personally identifiable information.
The proposed method utilizes the existing StyleGAN encoder~\cite{richardson2021encoding} to encode styles from encrypted images and the StyleGAN2 pretrained model~\cite{karras2020analyzing} to generate images.
We schedule a random key for encrypting images during the training process.

\subsection{Threat Model}
We consider a ciphertext-only attack (COA).
The goal of the attack is to recover personally identifiable visual information from encrypted face images.
We suppose an attacker knows the encryption (EtC) algorithm, but not the key.
We also assume a face dataset and a pre-trained StyleGAN2 generator are available to the attacker.
Any general face dataset is applicable and it is not necessary to have the same dataset as the one used by the StyleGAN2 model.
The attacker prepares encrypted images with random keys and has access to GPU-accelerated computing to train a StyleGAN encoder with the intent of recovering personal visual information from encrypted face images.

\subsection{StyleGAN Encoder Review}
The StyleGAN encoder is originally proposed for a general image-to-image translation task~\cite{richardson2021encoding}.
This encoder encodes images into the latent space of a pre-trained StyleGAN generator.
The encoder is trained by using a weighted combination of the pixel-wise $\mathcal{L}_2$ loss, the perceptual loss, the regularization loss for latent vectors, and the facial image recognition loss~\cite{richardson2021encoding}.
We refer interested readers to the original paper~\cite{richardson2021encoding}.
In our experiment, we did not use the regularization loss for latent vectors.

\subsection{Training Strategy}
We train the StyleGAN encoder~\cite{richardson2021encoding} by using pairs of plain images and encrypted ones.
Algorithm~\ref{algo:training} details the procedure of the training process.
The encoder $E$ encodes encrypted images to latent vectors that are combined with an average latent vector from the generator $G$.
Then, generator $G$ generates plausible images from the latent vectors.
During training, we schedule a random key every epoch, so that the encoder learns to extract styles from an encrypted image with any key.
Note that how often to change the key is important.
If we change the key too often, the encoder will not be able to learn the particular styles.
If we use only one key, the encoder will be biased toward the key.
Therefore, in our experiment, we changed the key every epoch and the results showed that the encoder $E$ is able to extract identifiable styles from the encrypted images.
\begin{algorithm}
\caption{StyleGAN Encoder Training for Proposed Attack\label{algo:training}}
\begin{algorithmic}[1]
   \State{Initialize encoder $E$}
   \State{Initialize and load pretrained StyleGAN2 generator $G$}
   \State{Initialize and load average latent code $\bar{w}$}
   \For{each epoch}
   \State{$K \leftarrow$ a random key} \Comment{Generate a random key.}
   \For{each plain image batch $\boldsymbol{I}$}
   \State{$\boldsymbol{I_e} \leftarrow \text{EtC}(\boldsymbol{I}, K)$} \Comment{Encrypt images.}
   \State{$\text{latent vectors} \leftarrow E(\boldsymbol{I_e})$} \Comment{Encode images to latent vectors.}
   \State{$\boldsymbol{I'} \leftarrow G(\text{latent vectors} + \bar{w})$} \Comment{Combine with an average latent code and decode the resulting latent code to an image.}
   \State{Calculate total loss $\mathcal{L}(\boldsymbol{I}, \boldsymbol{I_e}, \boldsymbol{I'})$}
   \State{Backpropagate and update the encoder $E$}
   \EndFor
   \EndFor
 \end{algorithmic}
\end{algorithm}

\subsection{Evaluation Metric\label{sec:lpips}}
We assess the perceptual quality of reconstructed images from the proposed attack by using the learned perceptual image patch similarity (LPIPS) metric~\cite{zhang2018unreasonable}.
This metric is well known and widely used as a perceptual loss in image generation tasks.
The previous attack also utilizes the LPIPS metric for evaluating the perceptual information of reconstructed images~\cite{madono2021gan}.
The higher LPIPS score means two images are further and the lower means two images are similar.
The LPIPS score between a plain image $I$ and an encrypted image $I_e$ is calculated by using extracted features from $L$ layers of a pretrained network as~\cite{zhang2018unreasonable};
\begin{equation}
d(I_e,I) = \sum_l \dfrac{1}{H_l W_l} \sum_{h,w} || w_l \odot ( \hat{y}_{ehw}^l - \hat{y}_{hw}^l ) ||_2^2,
\end{equation}
where $H_l$ and $W_l$ are a spatial dimension of a feature map, $w_l$ is a scaling vector, $\odot$ is an element-wise multiplication operation, and $\hat{y}_{hw}^l$ and $\hat{y}_{ehw}^l$ corresponding extracted feature maps of the $l$\textsuperscript{th} layer.

\section{Experiments and Discussion}
\subsection{Setup}
We used CelebA-HQ dataset~\cite{karras2018progressive}, which is a high-quality version of CelebA dataset~\cite{liu2015faceattributes}.
The dataset consists of 30,000 male and female face images, where 28,000 images were used for training and 2,000 images were reserved for testing.
We utilized the StyleGAN encoder code base from the original authors~\cite{richardson2021encoding} and a pre-trained StyleGAN2 model from~\cite{karras2020training}.
The StyleGAN2 was pre-trained on Flickr-Faces-HQ (FFHQ) dataset~\cite{karras2019style}.
We did not use the latent regularization loss and trained the encoder network for 500,000 steps.
A random key was substituted every epoch for encrypting images during the training as described in Algorithm~\ref{algo:training}.

For the ITN attack~\cite{ito2021image}, we used the available repository\footnote{https://github.com/ito-hiroki/learnable\_encryption\_robustness}, and for the GAN attack~\cite{madono2021gan} we utilized the released code\footnote{https://github.com/MADONOKOUKI/SIA-GAN} without an adaptation layer for simplicity.

\subsection{Results}
\begin{figure}[!t]
\centering
\includegraphics[width=\linewidth]{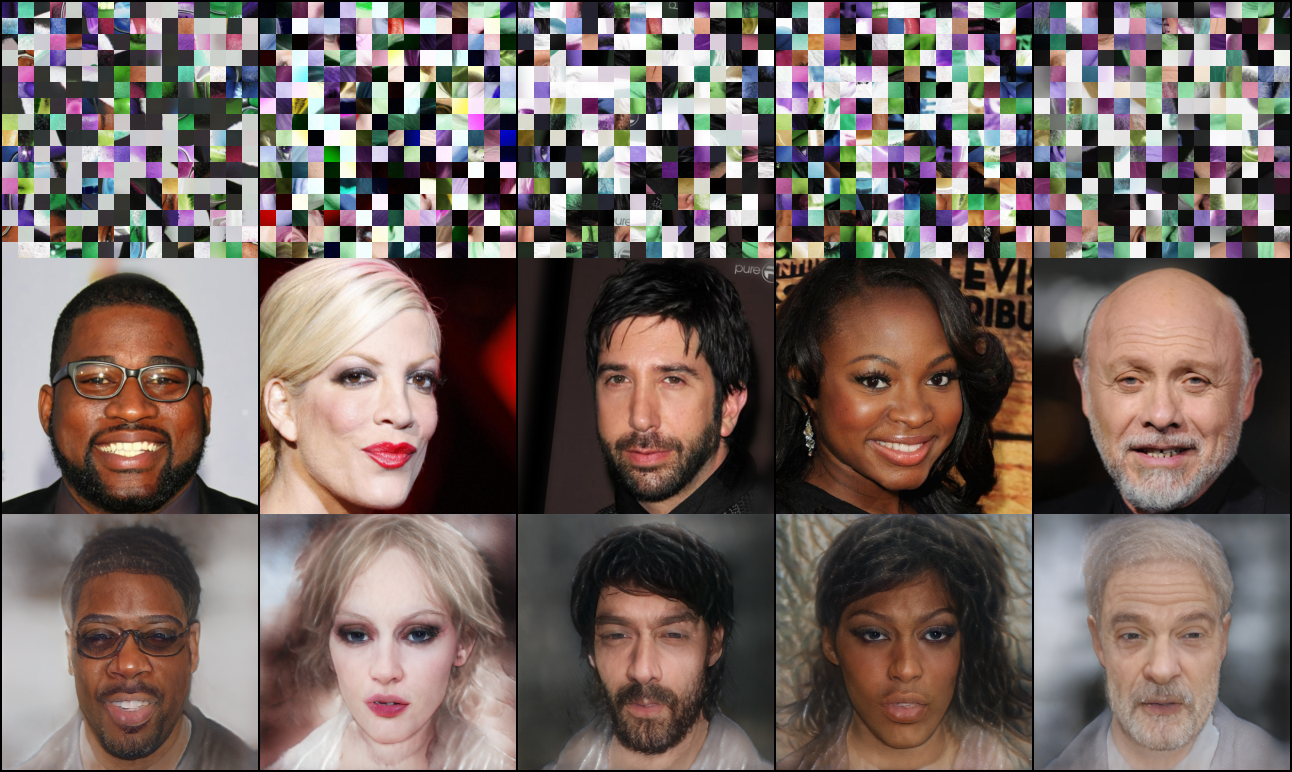}
\caption{Examples of encrypted images (first row), plain images (second row), and reconstructed images (third row).\label{fig:result}}
\end{figure}

We subjectively visualize the reconstructed images from encrypted images with different keys.
As shown in Fig.~\ref{fig:result}, the reconstructed images showed identifiable information such as skin color, gender, beard, eye glasses, etc.
We confirmed that encrypted images with different keys also revealed similar visual information.
Therefore, the proposed method is able to recover some identifiable information from encrypted images although the key is not known.

\subsection{Comparison}
We compared the proposed attack with state-of-the-art attack methods: ITN attack~\cite{ito2021image} and GAN attack~\cite{madono2021gan}.
As reported in~\cite{madono2021gan}, the previous attack cannot recover visual information when using a block scrambling encryption step.
EtC in~\cite{kurihara2015encryption} includes the block scrambling step and both ITN attack~\cite{ito2021image} and GAN attack~\cite{madono2021gan} cannot reconstruct images as shown in Fig.~\ref{fig:recon-examples}.
In contrast, the proposed method was able to reveal identifiable styles such as hair color, gender, etc.\ in Fig.~\ref{fig:recon-examples}.

\begin{figure}[!t]
\centering
\includegraphics[width=\linewidth]{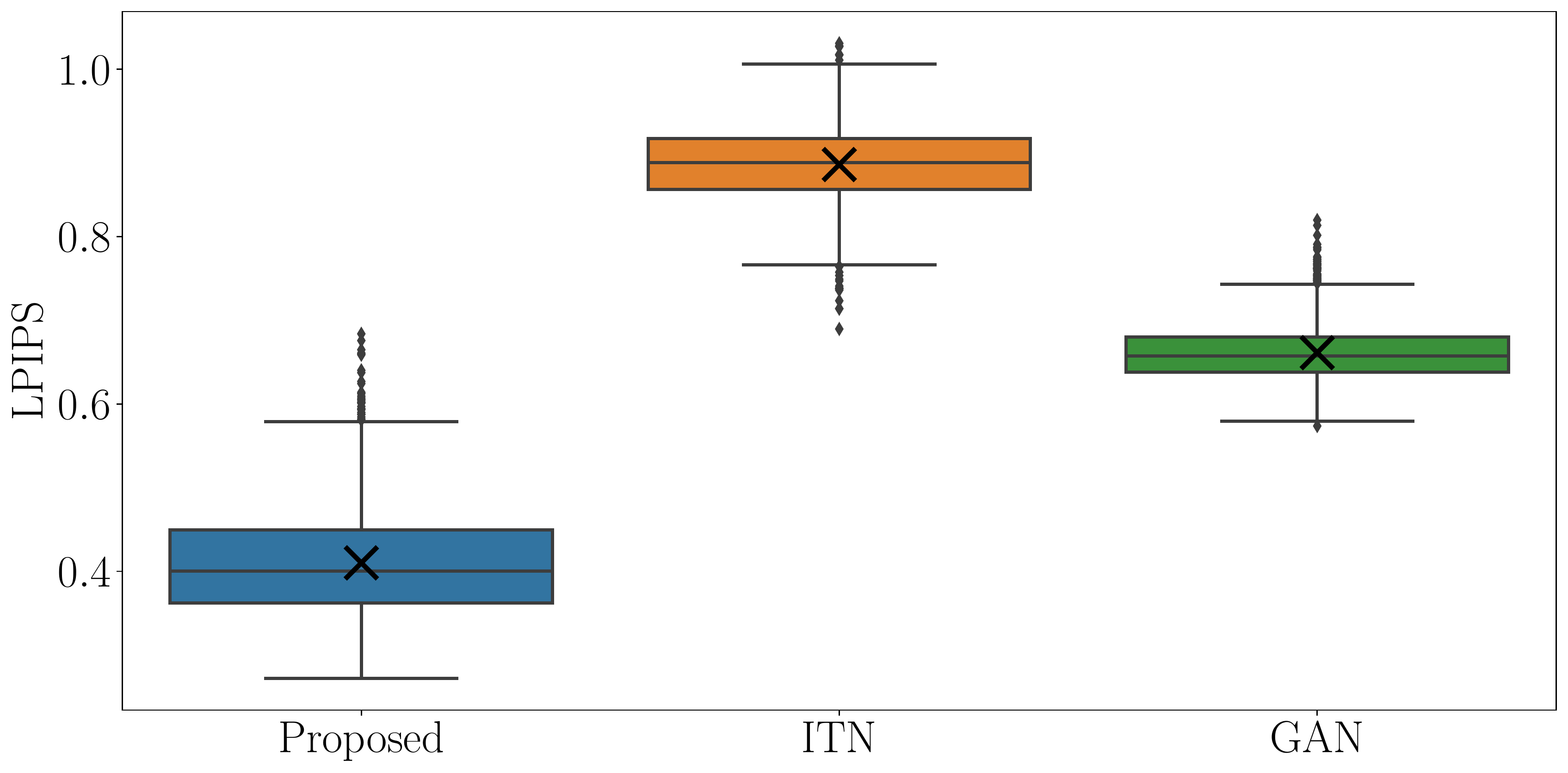}
\caption{LPIPS scores between plain images and reconstructed images for proposed attack, ITN attack~\cite{ito2021image}, and GAN attack~\cite{madono2021gan}. The scores were calculated over the validation set (2,000) images. Boxes span from first to third quartile, referred to as $Q_1$ and $Q_3$, and whiskers show maximum and minimum values in range of $[Q_1 - 1.5(Q_3 - Q_1), Q_3 + 1.5(Q_3 - Q_1)]$. Band and cross inside boxes indicate median and average values, respectively. Dots represent outliers.\label{fig:lpips}}
\end{figure}

Objectively, we also compared the proposed method with previous attacks in terms of LPIPS scores.
The LPIPS is the preferred way of measuring perceptual information between two image patches as described in Section~\ref{sec:lpips}.
We used 2,000 images from the validation set and results were plotted in Fig.~\ref{fig:lpips}.
The proposed method achieved smaller LPIPS scores meaning the resulting reconstructed images and plain images were perceptually similar.
In contrast, the reconstructed images from the ITN attack~\cite{ito2021image} and GAN attack~\cite{madono2021gan} were perceptually different, indicated by higher LPIPS scores.
Therefore, from the LPIPS  results, the proposed method outperformed the previous attack methods.

\subsection{Discussion and Limitation}
The proposed attack shows that it is possible to extract some styles from EtC images although the block scrambling step is included in the encryption.
Even though the encoder and decoder were trained on different datasets, experiment results proved that identifiable styles were recovered from the encrypted images regardless of the encryption key.

However, there are certain limitations.
\begin{itemize}
  \item The proposed method cannot reconstruct the identical face image as the plain image. It can only recover some facial features such as hair color, skin color, eye glasses, beard, etc.
  \item The proposed attack is only applicable to block scrambling-based EtC images.
    When pixel shuffling is used, the styles cannot be extracted by the encoder.
  \item The proposed attack relies on a pretrained StyleGAN2 generator which requires a great deal of resources to train.
    Without a high-quality StyleGAN2 generator, the proposed method cannot be performed.
\end{itemize}

Nevertheless, as generative models are getting better and better, existing perceptual image encryption methods should consider attacks from the generative models.
Experiment results in this paper show that although EtC images do not have perceptual information, some global styles are leaked.

\section{Conclusion}
We proposed an attack method to block scrambling-based EtC images by using an existing StyleGAN encoder and a pre-trained StyleGAN2 generator.
The encoder was trained by using plain and encrypted images with random keys scheduled every epoch.
As a result, the encoder is able to extract identifiable styles from the encrypted images with any key, and the pre-trained StyleGAN2 generator generates plausible images from the extracted styles encoded as latent vectors.
Experiment results show that the proposed method achieved smaller LPIPS scores compared to the state-of-the-art attack methods.
Although the proposed method reveals some identifiable information from face images, there are still some limitations.
As for future work, we shall further analyze the proposed attack to overcome the limitations.


\bibliographystyle{IEEEtran}
\bibliography{IEEEabrv,/Users/maung/Dropbox/refs}
\end{document}